\documentclass[a4paper,fleqn]{cas-dc}

\usepackage[numbers]{natbib}
\usepackage{graphicx}
\usepackage{capt-of} 
\def\tsc#1{\csdef{#1}{\textsc{\lowercase{#1}}\xspace}}
\tsc{WGM}
\tsc{QE}
\tsc{EP}
\tsc{PMS}
\tsc{BEC}
\tsc{DE}

\begin{document}
\let\WriteBookmarks\relax
\def\floatpagepagefraction{1}
\def\textpagefraction{.001}

\shorttitle{}


\title [mode = title]{Unifying Ontology Construction and Semantic Alignment for Deterministic Enterprise Reasoning at Scale}                      



%
\author[1,2]{Hongyin Zhu}[orcid=0000-0001-5786-7594]
\ead{zhuhongyin@yonyou.com}






\affiliation[1]{organization={Yonyou AI Lab}}





\affiliation[2]{organization={Yonyou Network Technology Co., Ltd.},
    country={}
    }






\begin{abstract}
While enterprises amass vast quantities of data, much of it remains chaotic and effectively dormant, preventing decision-making based on comprehensive information. Existing neuro-symbolic approaches rely on disjoint pipelines and struggle with error propagation. We introduce the large ontology model (LOM), a unified framework that seamlessly integrates ontology construction, semantic alignment, and logical reasoning into a single end-to-end architecture. LOM employs a construct-align-reason (CAR) pipeline, leveraging its unified architecture across all three stages: it first autonomously constructs a domain-specific ontological universe from raw data, then aligns neural generation with this structural reality using a graph-aware encoder and reinforcement learning, and finally executes deterministic reasoning over the constructed topology, node attributes and relation types. We evaluate LOM on a comprehensive benchmark constructed from diverse real-world enterprise datasets. Experimental results demonstrate that LOM-4B achieves 88.8\% accuracy in ontology completion and 94\% in complex graph reasoning tasks, significantly outperforming state-of-the-art LLMs. These findings validate that autonomous logical construction is essential for achieving deterministic, enterprise-grade intelligence.
\end{abstract}




\maketitle

\section{Introduction}
Contemporary enterprise AI faces a critical bottleneck in the prevalence of data silos that hinder the effective deployment of ontology intelligence at scale. Bridging the gap between academic advances and large-scale enterprise application remains a significant challenge, as real-world data is often fragmented and noisy. We address this by proposing an end-to-end framework that unifies ontology construction, semantic alignment, and deterministic reasoning within a single coherent architecture, paired with robust engineering practices for reliable deployment at scale.

Existing approaches often employ disparate model architectures to handle the distinct phases of construction, alignment, and reasoning, leading to significant engineering complexity and fragmentation. Furthermore, large language models exhibit impressive creativity, but often fail in deterministic reasoning tasks, particularly when applied to complex, noisy enterprise environments. We identify this limitation as a probabilistic wall, where scaling parameters alone yield diminishing returns for logical consistency. Standard LLMs frequently suffer from information loss during context graph transfer and remain constrained by a probabilistic wall that limits deterministic reasoning. To address these challenges, we unify these processes under a single cohesive model architecture. Critically, our approach to alignment extends beyond mere multimodal understanding to emphasize the dynamic capability of ontology updates, ensuring the model remains synchronized with evolving enterprise data.

A critical bottleneck lies in the dynamic evolution of the ontology—specifically, the non-trivial task of predicting latent relationships within sparse topological structures populated by entities with dense, unstructured textual attributes. In complex enterprise environments, the probabilistic nature of standard LLMs often fails to capture the strict deterministic dependencies required for valid ontology updates, leading to hallucinated links or missed connections. To overcome this, we leverage a graph-aware encoder \cite{zhu2024node} that explicitly models the topological context of each node, enabling the inference of missing relationships through a synergy of semantic content and structural patterns. This approach necessitates a rigorous text-ontology alignment training phase, where the model is optimized via a link prediction objective. By training to predict the existence and type of edges between nodes based on their textual descriptions and local subgraph structure, we force the model to ground its linguistic fluency in the logical reality of the graph. This alignment ensures that the inferred updates are not only semantically coherent but also topologically sound.

The second challenge involves adapting the LOM \cite{zhang2026construct} architecture for high-fidelity reasoning within the inherently noisy and ambiguous landscape of real-world enterprise data. In such environments, nodes are often laden with dense, heterogeneous textual attributes, and queries require a delicate balance between content-based filtering and structure-based deduction. A standard supervised learning approach often falls short in generalizing across this spectrum of tasks, struggling to distinguish subtle query nuances or to execute multi-hop reasoning chains without drifting. To this end, we incorporate reinforcement learning into our training pipeline. By defining a reward function that penalizes logical fallacies and rewards correct final answers and validated intermediate steps, we enable the model to autonomously discover optimal reasoning procedures (solution-step trajectories). This allows LOM to dynamically adapt its strategy—deeply integrating topological structure, node semantic content, and edge weights or types—to generate comprehensive solutions. Whether focusing on semantic matching for attribute-heavy queries or topological traversal for structural inquiries, LOM achieves robust performance across diverse tasks ranging from simple information retrieval to complex weighted graph algorithms like PageRank and minimum spanning tree.

Finally, the relentless velocity of AI advancement poses a strategic dilemma: engineering investments risk obsolescence almost as soon as they are deployed. To navigate this volatility and forecast the trajectory of intelligence, we present a theoretical analysis in section 4 that maps the evolution of AI onto a 10-Dimensional Cognitive Framework. This framework transcends linear scaling laws, proposing instead that breakthroughs occur through dimensional ascension—fundamental shifts in how information is structured and processed. We posit that current LLMs are hitting a 6D probabilistic wall, characterized by high-dimensional statistical correlation without grounding. By formalizing the criteria for ascending to the 7th dimension—where autonomous ontological construction precedes reasoning—we offer not just a retrospective taxonomy, but a predictive roadmap. This analysis provides a falsifiable theoretical basis for LOM's architecture and serves as a strategic guide for future research, distinguishing between transient engineering optimizations and enduring dimensional leaps.

We validate the effectiveness of our approach through comprehensive experiments on a benchmark constructed from real-world enterprise data, demonstrating significant improvements over existing baselines. The main contributions are as follows:

1. Real-world enterprise CAR methodology: Using authentic enterprise data at scale, we constructed an enterprise benchmark, trained LOM models, and conducted comprehensive evaluations—completing the full pipeline from benchmark construction to model training and assessment for deployment.

2. Full-stack ontology intelligence at scale: We complete the construct-align-reason pipeline over real enterprise data and surpass much larger models at every stage—88.8\% link prediction accuracy, high-fidelity semantic and structural alignment, and 95\% accuracy in deterministic reasoning across 19 diverse graph tasks.

3. 10-dimensional cognitive mapping: We locate contemporary techniques along a 1D–10D spectrum, characterize the 6D probabilistic wall, and formalize 7D logic autonomy where ontology construction precedes reasoning; this offers falsifiable criteria for dimensional ascension and practical guidance for building production systems beyond leaderboard metrics.

The remainder of this paper is organized as follows. Section 2 reviews related work in ontology construction, neuro-symbolic AI, and the development trends of agentic systems. Section 3 details the LOM architecture and the construct-align-reason pipeline. Section 4 presents the 10-Dimensional Cognitive Framework, providing a theoretical mapping for predicting technological evolutionary trends. Section 5 describes the experimental setup and results, demonstrating LOM's performance on enterprise benchmarks. Finally, Section 6 concludes the paper with a discussion of future directions.

\section{Related Work}
\subsection{Ontology Construction}
Recent advances in large language models have catalyzed a paradigm shift in ontology construction, moving from labor-intensive manual engineering toward automated or semi-automated pipelines. OntoEKG decomposes the ontology modeling task into an extraction module that identifies core classes and properties, and an entailment module that logically structures these elements into a hierarchy before serializing them into standard RDF \cite{oyewale2026llm}. Similarly, RIGOR combines three sources via retrieval-augmented generation—the database schema and its documentation, a repository of domain ontologies, and a growing core ontology—to prompt a generative LLM for producing successive, provenance-tagged delta-ontology fragments, with each fragment refined by a judge-LLM before integration \cite{nayyeri2025retrieval}. Beyond structured data sources, Le et al. introduce a structured, iterative methodology leveraging LLMs to optimize knowledge acquisition, automate ontology artifact generation, and enable continuous refinement cycles for ontological knowledge bases \cite{luyen2026development}. Cheung proposes Generative Ontology, which encodes domain knowledge as executable Pydantic schemas that constrain LLM generation via DSPy signatures, demonstrating that ontological structure can serve not only as a descriptive framework but also as a generative grammar for producing novel, structurally valid artifacts \cite{cheung2026generative}.

A complementary line of research investigates how the implicit knowledge within LLMs can be externalized into explicit symbolic representations through knowledge distillation. West et al. pioneer the Symbolic Knowledge Distillation framework, showing that careful prompt engineering and a separately trained critic model allow selective distillation of high-quality causal commonsense from GPT-3, producing a machine-authored commonsense knowledge graph \cite{zhu2023pre} that surpasses its human-authored counterpart in quantity, quality, and diversity \cite{west2022symbolic}. Acharya et al. provide a comprehensive survey of this area, concentrating on the process of distilling the intricate, often implicit knowledge within LLMs into more symbolic, explicit forms, and categorize existing techniques into direct distillation, multilevel distillation, and reinforcement-learning-based approaches \cite{acharya2024survey}. Liao et al. further advance this direction with NesyCD, which decouples general and specialized capabilities: general abilities are distilled into student neural networks, while specialized knowledge is extracted and stored within a symbolic knowledge base, enabling small language models to approach or even surpass the reasoning performance of models with significantly more parameters \cite{liao2025neural}. These symbolic distillation methods offer a viable pathway for constructing structured ontological resources from LLMs, complementing the direct ontology generation approaches described above.

\subsection{Neuro-Symbolic AI}
Neuro-symbolic AI seeks to unify the learning capabilities of neural networks with the formal reasoning power of symbolic systems, addressing the complementary weaknesses of each paradigm in isolation. This line of research studies the integration of symbolic reasoning and neural networks, and has been characterized along multiple dimensions including the approach to logical inference, the scope of learning, and the degree of coupling between symbolic and subsymbolic representations \cite{marra2024statistical}. On the purely symbolic side, Wolfram Alpha, built upon the Wolfram Language, represents a longstanding paradigm of knowledge-based computation. The Wolfram Language is a symbolic language deliberately designed with the breadth and unity needed to develop powerful programs quickly, representing everything—data, formulas, code, graphics, documents—as symbolic expressions \cite{wolfram2015elementary}. While this approach excels at precise, rule-governed computation over curated knowledge, it lacks the capacity for flexible learning from unstructured data—a limitation that motivates the hybrid neuro-symbolic paradigm. Existing neuro-symbolic architectures range from pipeline models that chain neural perception with symbolic reasoning, to hybrid architectures with tightly integrated components, to end-to-end differentiable models that embed logic directly in learning objectives \cite{sarker2022neuro}. Kautz's widely cited taxonomy further distinguishes integration patterns such as Symbolic[Neuro], Neuro→Symbolic, and Neuro[Symbolic], depending on whether the two components are composed serially, embedded, or tightly coupled \cite{kautz2022third}.

Among the most striking recent achievements in neuro-symbolic AI are DeepMind's AlphaGeometry and AlphaProof systems, both targeting mathematical reasoning at the olympiad level. AlphaGeometry is a neuro-symbolic system that uses a neural language model, trained from scratch on large-scale synthetic data, to guide a symbolic deduction engine through infinite branching points in challenging problems \cite{trinh2024alphageometry}. In this architecture, the language model provides creative intuition by predicting useful auxiliary constructions, while the symbolic engine carries out rigorous deductive steps—an interplay analogous to Kahneman's thinking, fast and slow. On a benchmark of 30 IMO geometry problems, AlphaGeometry solved 25, approaching the average score of human gold medalists, and its successor AlphaGeometry 2 further raised the solve rate to 84\% over all 2000–2024 IMO geometry problems \cite{liao2025neural}. Complementarily, AlphaProof is an AlphaZero-inspired agent that learns to find formal proofs through reinforcement learning by training on millions of auto-formalized problems, operating within the Lean theorem prover to guarantee the correctness of every logical step \cite{hubert2025olympiad}. At the 2024 IMO, AlphaProof solved three of five non-geometry problems, including the competition's hardest problem solved by only five human contestants. While these systems represent remarkable progress, they primarily employ neural networks to guide symbolic search or use symbolic verification to ground neural outputs—AlphaGeometry2 is a complex system with multiple modules, many of which are based on conventional mathematical software rather than on machine learning. Few existing systems achieve a true fusion in which the model autonomously constructs its own logical structures, highlighting an important open challenge for future neuro-symbolic research.

\subsection{State-of-the-Art in Agents}
The rapid evolution of LLM-based agents has given rise to increasingly sophisticated architectural patterns for managing complex, multi-step tasks. Wang et al. identify three fundamental architectural components that transform an LLM into an agent: a planning module, a memory module, and a tool-use module \cite{wang2024survey}. Building on this foundation, scaffolding techniques have emerged to decompose complex objectives into manageable sub-steps and guide agents through structured logical processes. MetaGPT introduces a meta-programming framework that incorporates efficient human workflows into LLM-based multi-agent collaborations, encoding Standardized Operating Procedures, SOPs, into prompt sequences to allow agents with human-like domain expertise to verify intermediate results and reduce errors \cite{hong2023metagpt}. The ReAct paradigm interleaves reasoning traces with action execution, enabling agents to dynamically plan, observe outcomes, and self-correct in a closed loop \cite{yao2022react}, while Reflexion extends this with explicit self-evaluation mechanisms that allow agents to learn from prior failures \cite{shinn2023reflexion}. To ensure reliability in high-stakes scenarios, human-in-the-loop architectures, HITL, inject expert oversight at critical decision points. Takerngsaksiri et al. introduce HULA, a human-in-the-loop LLM-based agent framework for software development that allows software engineers to refine and guide LLMs when generating coding plans and source code, demonstrating that such cockpit-style intervention capabilities can minimize overall development time while maintaining quality \cite{takerngsaksiri2025human}. More broadly, agentic AI workflows are structured around autonomous, task-specialized agents that coordinate via explicitly defined logical flows or graphs, with real-time monitoring dashboards providing operators with observability into agent reasoning chains and intervention hooks when execution deviates from expected trajectories \cite{zhang2024aflow}.

In the domain of AI-powered coding, autonomous agents have advanced from simple code-completion assistants to systems capable of handling end-to-end software development lifecycles. In March 2024, Devin AI released the first AI software engineer capable of autonomously completing end-to-end software tasks, achieving a 13.86\% resolve rate on SWE-bench \cite{yang2024swe}. SWE-agent further demonstrates that well-designed agent-computer interfaces significantly enhance an agent's ability to create and edit code files, navigate entire repositories, and execute tests, while OpenHands provides an open-source platform where coding agents can invoke shells, browsers, and external resources much like human developers \cite{wang2024openhands}. Complementing these developer-focused tools, OpenClaw introduces a hackable personal agent architecture that emphasizes persistent memory and dynamic skill acquisition, allowing the agent to evolve its capabilities through interaction and custom plugin integration \cite{Steinberger2025OpenClaw}. Multi-agent systems, where specialized agents collaborate on subtasks such as repository navigation, bug localization, patch generation, and verification, have evolved rapidly to address long-horizon challenges in software engineering. Despite these remarkable advances, it is important to note that current agentic innovations largely operate at the process management level—optimizing task decomposition, orchestration, and execution within existing logical frameworks—rather than autonomously constructing the underlying domain logic itself. The agents excel at following and refining prescribed workflows, but the conceptual structures and ontological schemas that govern their reasoning remain human-authored, highlighting an open frontier where agents could potentially learn to build their own formal knowledge representations.

\section{Approach}
We introduce the LOM \cite{zhang2026construct} which bridges the gap between probabilistic generation and deterministic reasoning through a unified construct-align-reason pipeline. Unlike traditional approaches that treat these as separate engineering tasks, LOM integrates them into a cohesive cognitive process. This section details how LOM autonomously builds its own logical universe in the construction phase, grounds abstract structures in semantic reality in the alignment phase, and executes rigorous algorithms within this self-created framework in the reasoning phase. Specifically, section 3.1 elaborates on the generative mechanisms for ontology construction from both unstructured and structured data. section 3.2 discusses the dual-encoder architecture for semantic-structural alignment and dynamic updates. section 3.3 demonstrates how LOM performs high-fidelity deterministic reasoning. Finally, section 3.4 outlines the two-stage training pipeline, including structural pre-training and reinforcement learning fine-tuning.

\subsection{Construct: Autonomous Ontology Synthesis}
The foundation of LOM is the ability to autonomously construct a high-fidelity ontology from structured and unstructured enterprise data. Unlike traditional methods that rely on rule-based extraction or manual curation, LOM employs a generative mechanism.

Ontology construction from unstructured data: This pipeline leverages a multi-stage LLM-driven architecture to transform unstructured enterprise text into a formal ontology. After initial element extraction using LOM with pydantic-enforced JSON schemas, the framework pivots to a logic-intensive hierarchy construction phase. Here, LOM is employed to perform semantic entailment, identifying latent subclass-superclass relationships. To ensure the structural integrity of the ontology, an iterative validation mechanism is applied, where the model systematically reasons through inheritance chains to verify their logical consistency and soundness. Finally, the validated conceptual model is serialized into the OWL turtle format, providing a robust, machine-readable knowledge foundation.

Ontology construction from structured data: This pipeline implements an iterative framework to systematically transform relational database schemas into formal OWL turtle ontologies. The process begins with a structured schema traversal, processing tables in an order determined by foreign key constraints to maintain relational integrity. For each element, a RAG module fetches semantic context from existing core ontologies and external documentation. LOM then synthesizes these inputs into delta ontology fragments, which undergo rigorous verification and optimization to ensure semantic consistency and syntactic compliance. This cycle repeats, incrementally merging validated fragments until all tables, columns, and relationships are exhaustively mapped into a logically sound global ontology.

The final phase of our approach employs a systematic four-stage process to merge disparate ontological structures into a unified OWL 2 DL model. First, concept alignment utilizes embedding similarity to match equivalent classes and properties, identifying semantic overlaps between the two source ontologies. Next, conflict resolution addresses naming collisions and hierarchical contradictions, leveraging LLMs to determine priority and merging strategies. During Hierarchy Integration, we establish cross-ontology relationships using rdfs:subClassOf and owl:equivalentClass axioms to create a cohesive taxonomic structure. Finally, the process concludes with Validation and Export, where reasoners ensure logical consistency and competency questions verify domain coverage before the final ontology is exported.

\subsection{Align: Semantic-Structural Fusion}
Our alignment mechanism fulfills two critical functions. First, it enables ontology understanding by statically grounding topological structures in real-world business semantics, ensuring that abstract graph nodes reflect concrete operational entities. Second, it facilitates ontology update, allowing for the dynamic adaptation of the graph structure in response to real-time data influxes, thereby maintaining synchronization with the evolving enterprise environment.

This capability for dynamic adaptation is particularly critical in real-time business environments, where the ontology must evolve instantaneously with the arrival of new data. Traditional batch-processing methods often fail to meet this requirement due to inherent latency. By extending the alignment mechanism to support incremental state changes, LOM allows the graph structure to be updated continuously based on the current ontology state. This ensures that the physical laws governing the information universe adapt in real-time to new empirical evidence, preserving the ontology's fidelity and relevance.

LOM conceptualizes the enterprise environment as a sparse graph, where nodes symbolize entities enriched with dense textual attributes and edges denote latent relationships. Through training on link prediction—specifically, identifying missing edges—LOM learns to deduce the underlying logical structure of the business domain. This mechanism effectively collapses the probabilistic noise inherent in raw data into a deterministic structural representation. In the context of our 10-Dimensional Cognitive Framework, this process signifies the transition from the 5th to the 6th dimension, as the model advances from perceiving a multiverse of potential connections to selecting the most logically consistent configuration.

Formally, we frame link prediction as a semantic-structural alignment task. Distinct from methods that serialize graphs into textual prompts, we employ a dedicated graph encoder $E_{\phi}$ to extract dense topological embeddings from the observed subgraph $\mathcal{G}_{obs}$. These embeddings are projected into the LOM latent space. The LOM then synthesizes this structural context with the semantic attributes of the target nodes $u$ and $v$ to predict the existence of a relationship:

$$ y = \text{LOM}_{\theta}(E_{\phi}(\mathcal{G}_{obs}), u, v) $$
where $y$ denotes the predicted relationship status, $E_{\phi}$ represents the graph encoder parameterizing the structural information, and $\text{LOM}_{\theta}$ serves as the reasoning core that integrates topological features with semantic understanding. Consequently, the predicted adjacency entry is defined by the indicator function:

$$ \hat{A}_{uv} = \mathbb{I}(y=\text{yes}) $$
where $\hat{A}_{uv}$ represents the predicted adjacency matrix entry for the edge between nodes $u$ and $v$, and $\mathbb{I}(\cdot)$ is the indicator function which equals 1 if the model output $y$ corresponds to a positive confirmation ("yes") and 0 otherwise.

This generative formulation distinguishes LOM from traditional embedding models that rely on static scalar scoring. By treating link prediction as context-aware reasoning, the model effectively decides on structural validity based on semantic evidence. Crucially, this inference capability supports the dynamic evolution of the ontology. As new data increments $\Delta D_t$ arrive, the graph state is recursively updated:

$$ \mathcal{G}_{t+1} = \text{Update}(\mathcal{G}_t, \Delta D_t; \theta_{align}) $$
where $\mathcal{G}_{t+1}$ and $\mathcal{G}_t$ denote the graph states at time steps $t+1$ and $t$ respectively. $\Delta D_t$ represents the incremental data influx, and $\text{Update}(\cdot)$ is the state transition function parameterized by the alignment model weights $\theta_{align}$. This mechanism ensures that LOM operates as a living system, continuously synchronizing its internal logical structure with the shifting reality of the enterprise environment.

\subsection{Reason: Logic Execution within Self-Constructed Axioms}
While LLMs demonstrate remarkable capability in simulating reasoning through probabilistic token prediction, they fundamentally lack a binding logical framework, often leading to hallucinations in complex scenarios. Despite Claude's remarkable heuristic reasoning \cite{knuth2026claude} on this particular directed Hamiltonian cycle problem, its ability to tackle diverse open questions and highly complex graph structures remains to be fully proven. The definitive 7D capability of LOM lies not in the reasoning process itself, but in its execution within a self-constructed axiomatic system. By treating the autonomously built ontology as a set of immutable physical laws, LOM transforms reasoning from a stochastic approximation into a rigorous, deterministic traversal. This ensures that every conclusion—whether derived via PageRank, minimum spanning tree, or shortest path algorithms—is mathematically guaranteed by the model's internal worldview, achieving a level of consistency unattainable by purely probabilistic architectures.

The reasoning process can be formalized as finding the optimal path or subgraph $\mathcal{G}^* \subseteq \mathcal{G}$ that satisfies a logical query $Q$. For a deterministic algorithm $\mathcal{A}$ (e.g., Dijkstra), the output is:

$$ \text{Result} = \mathcal{A}(\mathcal{G}_{aligned}, Q) $$

Unlike LLMs where $\text{Result} \sim P(\text{tokens} | Q)$, LOM ensures that the execution of $\mathcal{A}$ is bound by the constraints of the constructed ontology $\mathcal{G}_{aligned}$. This guarantees that if the ontology is correct, the reasoning result is mathematically certain.

\textbf{Beyond Retrieval: The Distinction from GraphRAG} It is crucial to distinguish LOM from GraphRAG. GraphRAG utilizes graphs as a retrieval index—organizing information topologically to help LLMs locate relevant text chunks. Its function is to enhance information access as external knowledge, but the final reasoning is still performed by the LLM's probabilistic next-token prediction engine. In contrast, LOM treats the ontology as logic laws, intrinsic intelligence. It does not merely read the graph; it executes deterministic algorithms on the graph structure itself. While GraphRAG allows an LLM to see more context, LOM provides the AI with a rigorous logical framework that strictly constrains its reasoning process.

\textbf{Convergent Evolution with AlphaGeometry} LOM shares a fundamental neuro-symbolic DNA with DeepMind's AlphaGeometry and AlphaProof. Both paradigms recognize that pure neural networks (LLMs) provide intuition but lack rigor, while pure symbolic systems provide rigor but lack scalability. AlphaGeometry combines a neural intuition engine with a symbolic deduction engine to solve mathematical proofs within a pre-existing axiomatic system. LOM applies this same fusion to the enterprise domain. However, LOM addresses a challenge of higher complexity: unlike mathematics, where axioms are pre-defined, enterprise axioms (business logic) are hidden within chaotic data. LOM must first autonomously construct the axiomatic universe (via the construct phase) before it can reason within it. Thus, LOM represents the same species of intrinsic intelligence evolution, adapted for the unstructured reality of the business world.

\subsection{Model Training}
Our training pipeline orchestrates a progressive evolution, beginning with semantic-structural alignment to ground the graph encoder within a deterministic ontology, and culminating in reinforcement learning optimization to fine-tune the logical reasoning capabilities of LOM-R1. In the foundational phase of semantic alignment and link prediction, we bridge the semantic-structural gap by aligning a pre-trained graph encoder \cite{zhu2025retracted} with the large language model. This alignment mechanism grounds abstract semantic representations within the concrete topological reality of the enterprise graph. Subsequently, we deploy a link prediction objective, optimized via cross-entropy, compelling the model to infer latent connections based on the synthesis of node attributes and existing graph structure. This process refines the graph encoder to capture deterministic dependencies, maximizing the likelihood of valid edges within the constructed ontology:
$$ \mathcal{L}_{construct} = \mathcal{L}_{CE}(\hat{A}, A) = - \sum_{(u,v) \in \mathcal{E}} \log P(A_{uv}=1 | \mathcal{G}) $$
where $\mathcal{L}_{construct}$ denotes the construction loss, $\hat{A}$ and $A$ represent the predicted and ground-truth adjacency matrices respectively, $\mathcal{E}$ is the set of edges in the training graph, and $P(A_{uv}=1 | \mathcal{G})$ signifies the probability of an edge existing between nodes $u$ and $v$ given the current graph state $\mathcal{G}$. This optimization ensures that the model first grounds semantic meaning before learning complex topological dependencies.

Following the structural grounding, we advance to logical reasoning optimization via reinforcement learning. To address the complexity of real-world queries—which often involve subtle attribute filtering and multi-hop weighted reasoning—we introduce reinforcement learning to fine-tune the LOM-R1 model. We employ the GSPO algorithm \cite{zheng2025group} for model training. Specifically, we utilize vLLM to generate n samples for each input query $Q$. These samples are evaluated using a reward reshaping function that scores both the final answer accuracy and the logical validity of the reasoning path.

The training objective is formulated as the maximization of the expected reward $J(\theta)$ across the distribution of generated reasoning trajectories:
$$ J(\theta) = \mathbb{E}_{y \sim \pi_\theta(\cdot|Q)} [R(y)] $$
where the reward function $R(y)$ is designed to balance result correctness with procedural soundness. It integrates a binary accuracy indicator with a continuous logical validity score, ensuring that the model is incentivized not only to retrieve the correct answer but to derive it through a rigorous inferential chain:
$$ R(y) = \alpha \cdot \mathbb{I}(y_{ans} = y_{gold}) + \beta \cdot \text{Score}_{logic}(y_{path}) $$
where $\mathbb{I}(\cdot)$ represents the indicator function, which equals 1 if the predicted answer $y_{ans}$ matches the gold standard $y_{gold}$, and 0 otherwise. $\text{Score}_{logic}(y_{path})$ quantifies the validity of the reasoning path, derived from the coherence of the graph traversal steps. The hyperparameters $\alpha$ and $\beta$ control the trade-off between answer accuracy and reasoning quality.

Furthermore, our optimization strategy incorporates a comparative analysis of logit distributions across diverse inference trajectories \cite{liu2024skywork}. This enables the model to discern and converge upon robust, deterministic logical patterns, effectively transcending the "probabilistic wall" inherent in traditional stochastic prediction.

\section{The 10-Dimensional Map of AI Evolution}
The theoretical foundation of our analysis is adapted from physicist Rob Bryanton's "Imagining the tenth dimension" \cite{bryanton2006imagining}, which describes the abstract structure of the universe across hierarchical levels. When this framework is mapped onto the evolutionary history of artificial intelligence, it reveals a startlingly accurate and predictive landscape of cognitive development.

In the current era of rapid AI proliferation, traditional metrics—such as parameter count or benchmark accuracy—often fail to capture qualitative leaps in intelligence. They measure linear growth within a single paradigm, for example "better next-token prediction", but cannot distinguish between a model that effectively retrieves information in 4D and one that autonomously structures knowledge in 7D. We posit that a dimensional framework is essential to:

1.  Map the cognitive territory: Precisely locate current technologies such as LLMs and agents on an evolutionary spectrum rather than a flat leaderboard.

2.  Identify the probabilistic wall: Reveal why scaling laws are hitting a ceiling at the 6th dimension, where models can navigate possibilities but cannot alter the underlying logical rules.

3.  Predict the trajectory: Clarify that the path to AGI requires a dimensional ascension—mastering the ability to autonomously construct the logical "physics" of a problem—rather than merely accelerating within existing dimensions.

We contextualize existing AI paradigms within this framework below.

\begin{table*}[htbp]
\centering
\caption{A 10-Dimensional Map of AI Evolution}
\label{tab:ai_evolution}
\begin{tabular}{lllp{6cm}} 
\toprule
\textbf{Dimension} & \textbf{Spatial Concept} & \textbf{AI Evolutionary Stage} & \textbf{Representative Form} \\ 
\midrule
1D-3D & Line/Plane/Volume & Statistical Learning & Rule-based Systems, Machine Learning, Deep Learning \\
4D & Time, Causal Sequences & LLMs & Transformers \\
5D & All Possible Timelines & Basic Agents & Planning, Hypothesis Reasoning \\
6D & Jumping Between Timelines & Meta-Cognitive Agents & Self-Reflection, Tool/Skill Building \\
7D & Different Physical Constants & LOM & Autonomous Logic Reconstruction \\
8D-10D & Multiverse/Omniscience & ASI & Paradigm Discovery, Singularity \\
\bottomrule
\end{tabular}
\end{table*}

\subsection{From 1D to 4D: The Evolution of Knowledge Representation}
The journey of AI can be mapped as a progressive conquest of spatial dimensions, representing increasing degrees of freedom in data representation.

1D, line: rule-based systems. Early AI operated on linear if-then logic. Like a point moving along a single line, these systems could only follow pre-defined paths. They were deterministic but rigid, unable to handle any deviation from the hard-coded rules.

2D, plane: traditional machine learning. With the advent of statistical learning such as SVMs and random forests, AI gained the ability to optimize decision boundaries across a two-dimensional plane of features. This allowed for generalization beyond exact matches, but representation remained flat and feature-dependent.

3D, volume: deep learning. Neural networks introduced depth through layers, allowing models to learn hierarchical representations of data. This three-dimensional understanding enabled the capture of complex, non-linear patterns in static inputs like images or fixed vectors.

4D, time: large language models. The transformer architecture introduced the fourth dimension, time as sequence. LLMs do not just process static data; they understand the causal flow of tokens, predicting the future $t+1$ based on the history $t-n$. This mastery of sequential probability allows them to generate fluent, context-aware text. However, their reasoning is fundamentally temporal—a probabilistic continuation of a sequence—rather than spatial or structural. When strictly causal chains are broken or when circular logic is required, as in many graph problems, they often hallucinate because they lack a stable, time-independent logical structure.

\subsection{From 5D to 6D: The Ceiling of Meta-Cognition}
The concept of metacognition \cite{flavell2024metacognitive}, originally proposed in 1976 by developmental psychologist John Flavell, describes the capability of thinking about thinking—monitoring and regulating one's own cognitive processes. In our framework, this distinguishes 6D intelligence from 5D.

5D, all possible timelines: basic agents. Systems like AutoGPT or BabyAGI operate in the fifth dimension. They can generate multiple plans, often called timelines, to solve a task and select the optimal decision path based on predictive outcomes. However, they lack the ability to step outside their current execution path to evaluate its validity against a higher-order framework.

6D, jumping between timelines: meta-cognitive agents. Advanced agents exhibit 6D behavior by not just executing tasks, but creating the means to execute them. This includes: (1) Scaffolding: decomposing complex goals into structured sub-processes to guide reasoning, such as OpenClaw \cite{Steinberger2025OpenClaw}. (2) Cockpits: interfaces for real-time monitoring and intervention, such as HULA \cite{takerngsaksiri2025human}, allowing for dynamic strategy adjustment. (3) Agentic code reasoning: instead of directly answering a query, systems like Devin, Glean, SWE-agent \cite{yang2024swe} write and execute code as tools to derive the answer.

This shift—from directly solving requirements to creating tools that solve requirements, and ultimately to building systems that mass-produce such tools—marks the peak of current AI capabilities. Yet, it hits the "probabilistic wall." Even a 6D agent is ultimately a probabilistic engine wrapping another probabilistic engine. It can optimize how it navigates the search space, but it cannot guarantee the correctness of the underlying logical laws; for example, it might write code that is syntactically correct but logically flawed due to a misunderstanding of the domain ontology. It cannot "rewrite the physics" of the problem; it can only build better vehicles to traverse it.

\subsection{The 7D Breakthrough: LOM's Logic Autonomy}
In the 7th dimension, we deal with different physical constants or, in the AI context, different logical frameworks. A 6D agent can explore all possibilities within a given world, for example a specific game or business process, but it cannot rewrite the rules of that world. 

LOM operates in the 7th dimension because it autonomously constructs the universe, the ontology, in which it reasons. By defining the entities \cite{zhu2019flexner}, relations \cite{zhu2022switchnet}, and constraints, the physical constants of the enterprise data, LOM creates a deterministic environment where logic is not just probable, but structural and binding. Just as our physical universe follows laws like Newton's First Law, every enterprise operates on its own set of intrinsic axioms—business rules, regulatory constraints, and operational workflows. However, unlike physics, these enterprise axioms are rarely summarized into clear mathematical formulas; they are hidden within chaotic data and tacit knowledge. The capability of autonomous ontology construction is essentially the process of exploring and formalizing these hidden axioms, transforming an opaque enterprise into a computable entity. This capability allows LOM to solve problems that are impossible for 4D, 5D, and 6D models, such as finding the exact shortest path in a graph or strictly adhering to a complex supply chain policy.

Fundamentally, LOM's CAR pipeline is not merely a sequence of steps but a unified cognitive mechanism that embodies 7D intelligence.

Integrated understanding and reasoning: Enterprise data is a duality of unstructured text, which requires semantic understanding, and structural relationships, which require logical reasoning. Traditional pipeline systems treat these separately—calling an NLP model for text and a graph model for structure—resulting in information loss and low efficiency. LOM integrates these via the Align phase, ensuring that every logical deduction in the Reason phase is grounded in the full semantic context of the node attributes.

End-to-end internalization: By autonomously constructing the ontology in the construct phase and grounding it semantically in the Align phase, LOM internalizes the entire problem space. It does not need to repeatedly query external modules or databases; it thinks within a self-contained, consistent universe. This internalization is what allows for deep, multi-hop reasoning that considers all available information simultaneously, rather than fragmenting the problem into isolated sub-tasks.

Efficiency over assembly: Unlike agentic workflows that rely on assembling disparate tools, which is slow and error-prone, LOM's end-to-end capability enables it to execute complex reasoning with the speed and coherence of a single model, overcoming the efficiency bottlenecks of piecemeal system integration.

\subsection{Reflection: Acceleration vs. Ascension}

This mapping compels us to reflect on the current trajectory of AI development from a 10-dimensional perspective.

Acceleration within 4D: Most efforts in scaling LLMs, more parameters and more data, are essentially accelerating within the fourth dimension. They are building a faster, more fluent predictor of the next token, mastering the probability of sequences.

Navigation within 5D and 6D: Agentic workflows such as CoT and ReAct and tool-use frameworks are navigating the fifth and sixth dimensions more efficiently. They allow systems to explore multiple plans and create tools to traverse the problem space. However, they remain trapped within the probabilistic nature of the base model, the probabilistic wall, unable to guarantee the logical validity of the universe they operate in.

Ascension to 7D: LOM represents a dimensional ascension. By autonomously constructing the ontology and internalizing the problem space through the CAR pipeline, LOM builds the internal combustion engine, a 7D logic engine, rather than a faster horse, that is, better 4D and 5D models. It defines the physical constants of the enterprise, transforming chaotic data into a deterministic, computable entity.

This analysis suggests a critical distinction for future research: optimizing performance within current paradigms versus enabling structural shifts to higher-dimensional reasoning. We argue that achieving robust general intelligence necessitates moving beyond data scaling alone, requiring architectures capable of autonomous logical construction.
\section{Experiments}

\subsection{Dataset}
Our benchmark dataset is collected directly from internal production systems. These datasets encompass comprehensive business scenarios including human resources, finance, assets, production, supply chain, and sales—HR, finance, supply chain, and manufacturing—integrating both structured databases and unstructured documentation. This shift from synthetic to real-world data was pivotal.

Our benchmark dataset consists of 19 diverse graph reasoning tasks, designed to test both structural and semantic understanding. For each core reasoning task, we constructed a balanced dataset comprising 500 training samples and 100 testing samples. Additionally, for the link prediction task in ontology construction, we utilized a significantly larger dataset consisting of 8,500 real-world business graphs for training and 1,000 graphs for testing. These samples were derived from real-world enterprise scenarios, ensuring the model's performance reflects practical applicability rather than theoretical benchmarks.

\subsection{Experimental Setup}
We trained the LOM-4B, LOM-32B, and LOM-Construct models on a distributed computing cluster with a global batch size of 64. The training infrastructure comprised four servers, each equipped with 8 × NVIDIA A800 80GB GPUs, 2 × Intel Xeon CPUs, and 1TB RAM, totaling 32 GPUs. All models were trained with full precision, FP32, to ensure maximum fidelity in capturing the subtle nuances of the text-attributed graphs.

\subsection{Evaluation Metrics}
The primary metric for all tasks is accuracy, defined as the percentage of test samples where the model's output exactly matches the ground truth. 

\subsection{Ontology Completion Performance}
We first validated the model's capacity for autonomous ontology completion. On this challenging test set, LOM achieved an accuracy of 88.8\% in link prediction. It is important to note that unlike clean academic benchmarks, these graphs are derived from raw enterprise data containing significant noise and unstructured text attributes. Achieving such high fidelity in this context is crucial; it ensures that the "universe" in which the model reasons is a faithful representation of the underlying business logic, rather than a hallucinated structure. This step effectively filters out the noise inherent in the raw data, providing a solid 6D foundation for the subsequent 7D reasoning tasks.

\begin{center}
\includegraphics[width=0.45\textwidth]{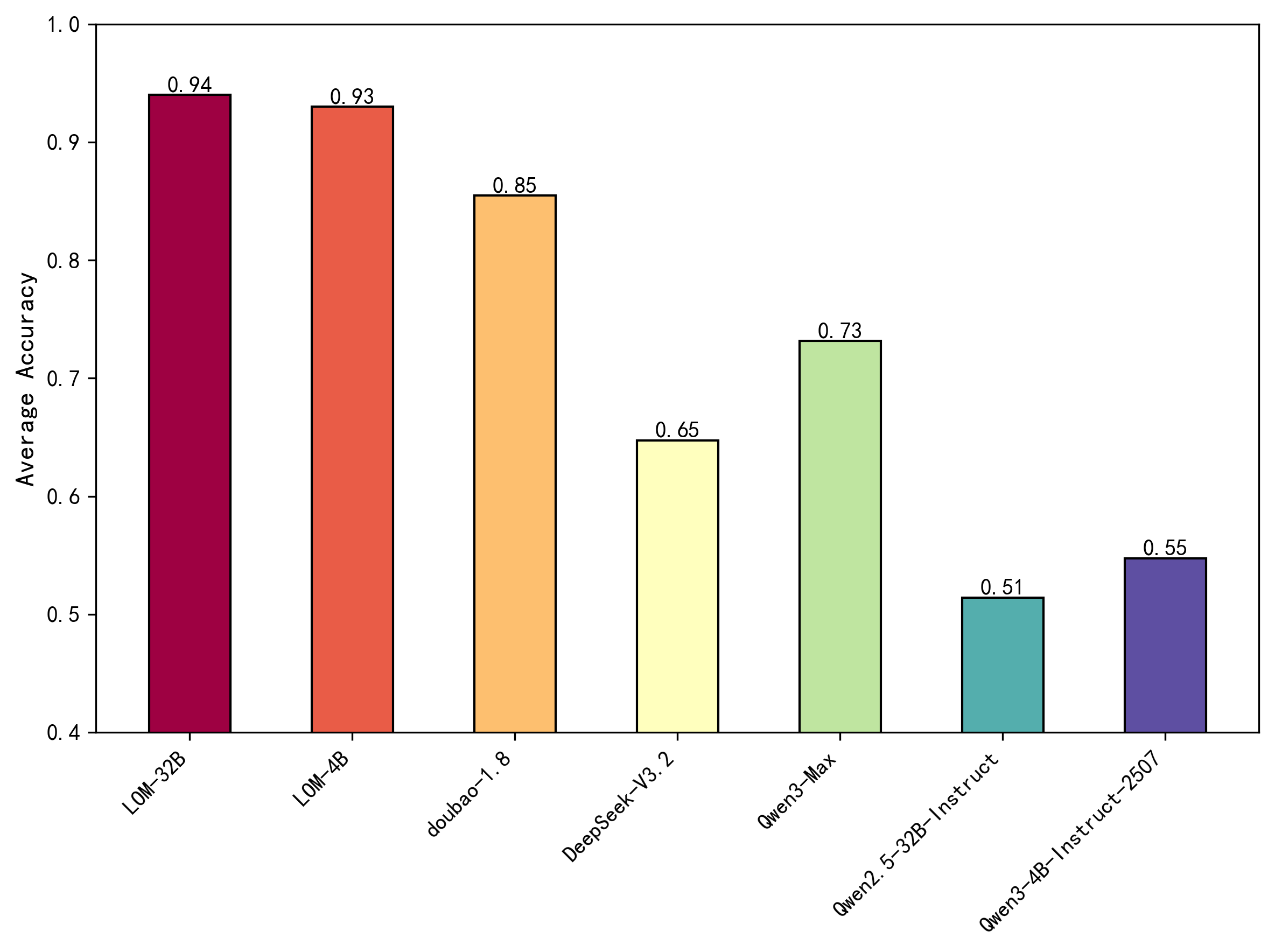}
\captionof{figure}{Comparative Accuracy on Enterprise Ontology Tasks}
\label{fig.compare}
\end{center}

\begin{figure*}[t]
\centering
\includegraphics[width=0.8\textwidth]{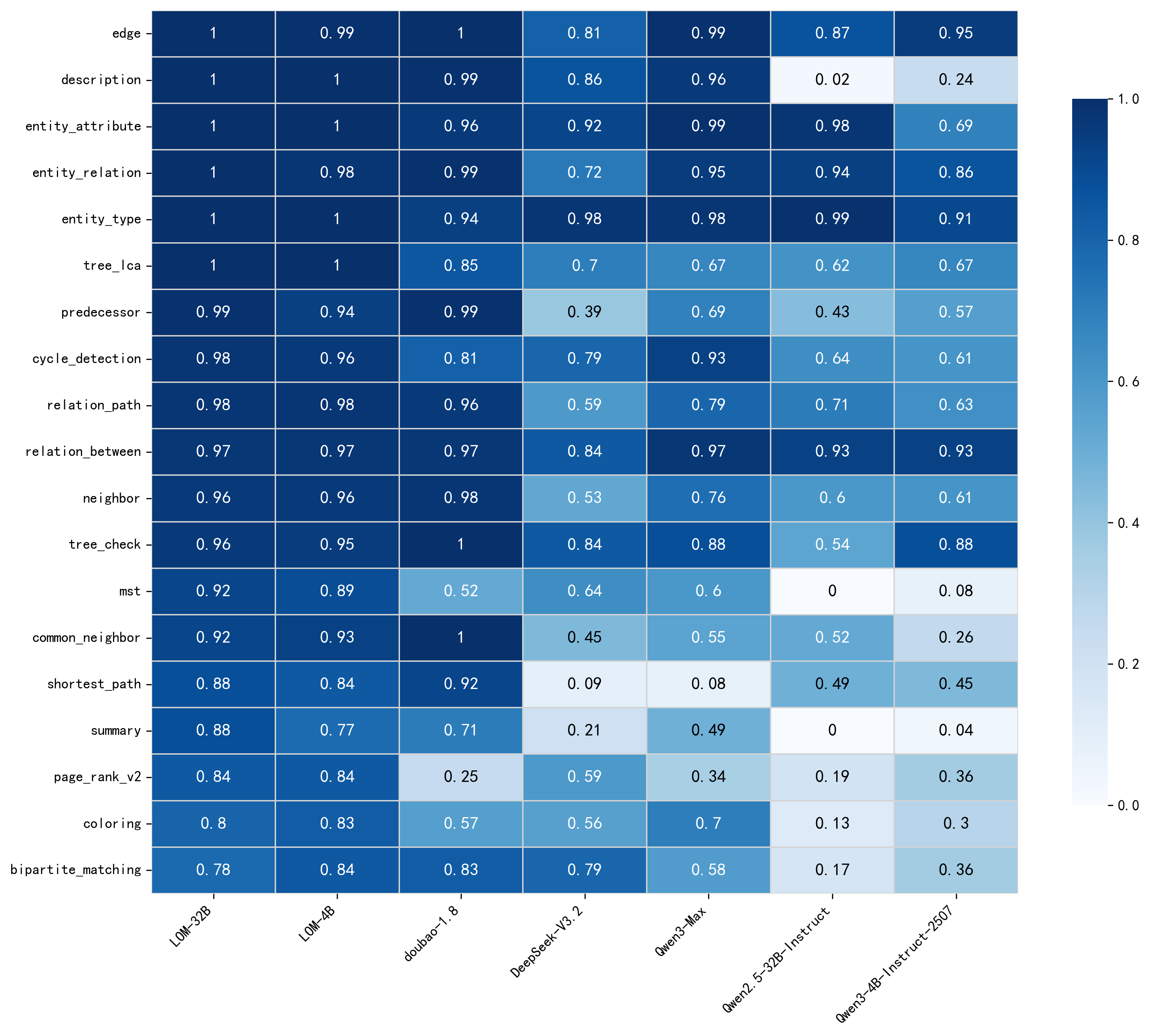}
\captionof{figure}{Heatmap of Per-Task Accuracy Across Models}
\label{fig.2}
\end{figure*}
\subsection{Graph Reasoning Performance}
We conducted a comprehensive evaluation across 19 graph reasoning tasks, comparing LOM-4B and LOM-32B against large-scale LLMs, including Doubao-1.8, Qwen3-Max, and DeepSeek-V3.2. As shown in Figure \ref{fig.compare}, LOM delivers consistently strong performance across topology, path/structure, algorithmic, and semantic tasks, achieving average accuracies of 0.93 (LOM-4B) and 0.94 (LOM-32B). Scaling from 4B to 32B improves several challenging tasks while preserving structural reliability; for example, Shortest Path increases from 0.84 to 0.88, MST from 0.89 to 0.92, Cycle Detection from 0.96 to 0.98, and Summary from 0.77 to 0.88, as shown in Figure \ref{fig.2}.

In contrast, general LLMs exhibit a pronounced bifurcation: they can remain competitive on surface-level semantic tasks—for example, Doubao-1.8 achieves 0.99 on description—yet collapse on tasks requiring deterministic structural constraints. On shortest\_path, Qwen3-Max scores 0.08 and DeepSeek-V3.2 scores 0.09, and on MST, Qwen2.5-32B scores 0. This gap highlights the necessity of 7D-style logic execution within an internal ontology, rather than purely probabilistic next-token prediction.

\subsection{Analysis and Discussion}

\noindent\textbf{Unified Semantics and Structure.}
The results validate the CAR architecture. Unlike pipeline systems that fragment semantic parsing and graph reasoning, LOM internalizes both end-to-end. This integration allows LOM to maintain state-of-the-art performance on semantic tasks (e.g., description 1.00) while excelling at structure-sensitive tasks (e.g., relation path 0.98), where traditional models often falter. The align phase ensures semantic outputs are grounded in the same rigorous ontology used for reasoning, preventing the trade-off between fluency and structural accuracy.

\noindent\textbf{Overcoming the Probabilistic Wall.}
A clear boundary emerges in algorithmic tasks requiring strict constraint satisfaction. LOM achieves high accuracy on MST (0.92) and shortest path (0.88), whereas general LLMs collapse (e.g., Qwen2.5-32B scores 0 on MST). This confirms the 7D claim: probabilistic scaling alone cannot master deterministic logic. Without an internal binding structure, even large models fail to preserve multi-step invariants, illustrating a probabilistic wall that parameter scaling does not breach.

\noindent\textbf{Logic Density over Parameter Scale.}
LOM demonstrates that logic density outweighs parameter count for complex reasoning. LOM-4B (93\% average accuracy) and LOM-32B (94\%) significantly outperform models 10x their size (e.g., Qwen2.5-32B, 70B+ models). By constraining the model with strict ontological rules (logic laws), LOM achieves higher cognitive density, proving that neuro-symbolic integration is a more efficient path to reliable reasoning than unsupervised pre-training alone.

\section{Conclusion}
In this paper, we introduced the LOM, a neuro-symbolic framework that unifies ontology construction, semantic alignment, and logical reasoning to address the inherent limitations of probabilistic models in deterministic domains. Through the proposed CAR pipeline, LOM autonomously formalizes latent knowledge into explicit ontological structures, effectively overcoming the probabilistic wall characterized in our 10-Dimensional cognitive framework. Empirical evaluations on large-scale enterprise benchmarks demonstrate that LOM not only achieves superior performance on complex reasoning tasks compared to models with significantly larger parameter counts but also ensures logical consistency and interpretability. Our findings indicate that equipping language models with the capability for autonomous logic construction—7D logic autonomy—is a critical direction for advancing the reliability and utility of AI in high-stakes enterprise applications.

Building upon this 7D logic foundation, our future work will extend LOM's capabilities into autonomous action execution, strategic decision-making, and dynamic skill acquisition. This evolution will transition LOM from a passive reasoning engine to an active agent capable of intervening in real-world enterprise processes, effectively closing the loop between insight and impact.

\printcredits

\bibliographystyle{cas-model2-names}

\bibliography{cas-refs}

\end{document}